\documentclass[12pt]{article}
\usepackage[utf8]{inputenc}
\usepackage[T1]{fontenc}
\usepackage{lmodern}
\usepackage{geometry}
\usepackage{amsmath,amssymb,amsthm}
\usepackage{authblk}
\usepackage{graphicx}
\usepackage{booktabs}
\usepackage{multirow}
\usepackage{float}        
\usepackage[section]{placeins} 
\usepackage{hyperref}
\hypersetup{colorlinks=true,linkcolor=blue,citecolor=blue,urlcolor=blue}

\title{Wasserstein Regression as a Variational Approximation of Probabilistic Trajectories through the Bernstein Basis}
\author[1]{Maksim I. Maslov}
\author[1]{Alexander V. Kugaevskikh}
\author[1]{Matthew S. Ivanov}
\affil[1]{ITMO University, Saint Petersburg, Russia}
\date{October 30, 2025}

\begin{document}
\maketitle

\begin{abstract}
This paper considers the problem of regression over distributions, which is becoming increasingly important in machine learning. Existing approaches often ignore the geometry of the probability space or are computationally expensive. To overcome these limitations, a new method is proposed that combines the parameterization of probability trajectories using a Bernstein basis and the minimization of the Wasserstein distance between distributions. The key idea is to model a conditional distribution as a smooth probability trajectory defined by a weighted sum of Gaussian components whose parameters---the mean and covariance---are functions of the input variable constructed using Bernstein polynomials. The loss function is the averaged squared Wasserstein distance between the predicted Gaussian distributions and the empirical data, which takes into account the geometry of the distributions. An autodiff-based optimization method is used to train the model. Experiments on synthetic datasets that include complex trajectories demonstrated that the proposed method provides competitive approximation quality in terms of the Wasserstein distance, Energy Distance, and RMSE metrics, especially in cases of pronounced nonlinearity. The model demonstrates trajectory smoothness that is better than or comparable to alternatives and robustness to changes in data structure, while maintaining high interpretability due to explicit parameterization via control points. The developed approach represents a balanced solution that combines geometric accuracy, computational practicality, and interpretability. Prospects for further research include extending the method to non-Gaussian distributions, applying entropy regularization to speed up computations, and adapting the approach to working with high-dimensional data for approximating surfaces and more complex structures.
\end{abstract}

\textbf{Keywords:} Wasserstein regression, Bernstein basis, optimal transport, Gaussian mixtures, distribution regression.

\footnote{ The research was carried out within the state assignment of Ministry of Science and Higher Education (project FSER-2025-0004).}

\pagebreak
\section{Introduction}
Modern machine learning increasingly requires moving from point predictions to modeling entire distributions, which is especially important in areas where uncertainty estimation matters~\cite{chen2023wasserstein}. Classical regression methods are insufficient when the target variable is a random variable with a complex dependence structure on input features, which has led to the active development of distribution regression~\cite{kneib2023rage}.

Existing approaches either ignore the geometry of the space of probability measures, are computationally expensive, or require large amounts of data for reliable density estimation~\cite{belbasi2025its,zhou2023wasserstein}. This work proposes a Wasserstein-regression method that combines parameterization based on the Bernstein basis with geometrically motivated minimization of the Wasserstein distance. The main goal is to construct a smooth probabilistic trajectory that approximates the conditional distribution of the response.

The novelty lies in the synthesizing ofzing of ideas of variational approximoptimal transport geometry, and parametric modeling. The proposed method provides interpretability through explicit control points, computational efficiency, and robustness on complex nonlinear dependencies, as confirmed by experimental studies.

\section{Related Work}
\paragraph{Distribution regression.} Modeling complete probability distributions instead of scalar responses has become increasingly popular, leading to the field of distribution regression~\cite{kneib2023rage}. Classical approaches (based on embedding distributions in RKHS using MMD kernels) allow standard linear models in feature space, but they ignore the geometry of probability measures, which reduces the effectiveness for complex data~\cite{cherfaoui2022discrete}.

A more natural approach is Wasserstein regression~\cite{chen2023wasserstein}. It minimizes the Wasserstein distance between the predicted distribution and the true one $P(y\mid x)$, thereby accounting for the shape and displacement of mass between them. The geometric structure makes the model more robust to noise and outliers. Wasserstein regression has been applied to stochastic dependencies and functional data~\cite{chen2023wasserstein}, and extended to generative models (Wasserstein Generative Regression)~\cite{song2023wasserstein}, where a neural approximation of the conditional distribution is used.

However, the main drawback is the computational complexity due to optimal transport. Entropic regularization (the Sinkhorn method)~\cite{meunier2022distribution} and simplified variants -- sliced Wasserstein metrics that approximate multidimensional distance with one-dimensional projections~\cite{nguyen2025fast} -- are common remedies.

Another approach is conditional density estimation (CDE), where for each input $x$ one builds a density estimate $p(y\mid x)$, e.g., as in the LinCDE model~\cite{gao2022lincde}. This provides more complete information than a simple mean but is harder to tune and requires larger data volumes.

More recent work includes distributionally robust regression. Liu et al.\ consider nonparametric regression within a Wasserstein ball of distributions, obtaining nontrivial upper risk bounds under model shifts~\cite{liu2025wasserstein}. These methods consider worst cases in a neighborhood of the sample in the Wasserstein sense. Curves in the space of measures have also been parameterized via B-splines with geodesic averaging in Wasserstein space (Wasserstein Lane--Riesenfeld)~\cite{banerjee2025efficient}, emphasizing preservation of geometry along trajectories using subdivision schemes and segment-wise averaging.

Overall, distribution regression balances expressiveness (full distributional forms) and computational/statistical complexity (data requirements and optimization complexity).

\section{Method}
Let observations be pairs $(x_i,y_i)$, $i=1,\dots,n$, where $x_i\in\mathbb{R}$ is a scalar input and $y_i\in\mathbb{R}^d$ is a vector-valued response (trajectory coordinates in $d$-dimensional space). At each position $x_i$, the true response is a probability measure $P(y\mid x_i)$. The goal is to build a parametric model $P_\theta(y\mid x)$ that approximates the family $\{P(\cdot\mid x)\}$ under the $p$-Wasserstein metric with $p=2$ (denoted $W_2$).

\subsection*{Bernstein parameterization of the curve}
Let $N\in\mathbb{N}$ be the polynomial degree. The degree-$N$ Bernstein basis polynomials are
\begin{equation}
b_{i,N}(t)=\binom{N}{i}(1-t)^{N-i}t^{i},\qquad i=0,\dots,N,\quad t\in[0,1].
\end{equation}
For mixture component $k$ we define control means $\{\mu_{k,i}\in\mathbb{R}^d\}_{i=0}^{N}$ and symmetric positive semidefinite control covariances $\{\Sigma_{k,i}\in\mathbb{R}^{d\times d}\}_{i=0}^{N}$. Then the component mean and covariance at $t$ are
\begin{align}
\mu_k(t) &= \sum_{i=0}^{N} b_{i,N}(t)\,\mu_{k,i},\\
\Sigma_k(t) &= \sum_{i=0}^{N} b_{i,N}(t)\,\Sigma_{k,i}.
\end{align}
The component distribution is Gaussian
\begin{equation}
P(y\mid t,k)=\mathcal{N}\!\big(\mu_k(t),\,\Sigma_k(t)\big).
\end{equation}
The overall model is a mixture of $K$ components with weights $w_k\ge 0$, $\sum_{k=1}^{K}w_k=1$:
\begin{equation}
P(y\mid t)=\sum_{k=1}^{K} w_k\, \mathcal{N}\!\big(\mu_k(t),\,\Sigma_k(t)\big).
\end{equation}

\subsection*{Squared $W_2$ between Gaussians}
For $\mathcal{N}(\mu_1,\Sigma_1)$ and $\mathcal{N}(\mu_2,\Sigma_2)$, the squared $W_2$ distance is
\begin{equation}
\label{eq:w2}
W_2^2\!\big(\mathcal{N}(\mu_1,\Sigma_1),\mathcal{N}(\mu_2,\Sigma_2)\big)
= \|\mu_1-\mu_2\|_2^2 + \mathrm{Tr}\!\Big(\Sigma_1+\Sigma_2 - 2\big(\Sigma_2^{1/2}\Sigma_1\Sigma_2^{1/2}\big)^{1/2}\Big),
\end{equation}
where $(\cdot)^{1/2}$ denotes the symmetric positive square root and $\mathrm{Tr}$ is the matrix trace.

\subsection*{Empirical risk and training}
Normalize inputs by
\begin{equation}
t_i=\frac{x_i-\min_j x_j}{\max_j x_j - \min_j x_j},\qquad i=1,\dots,n.
\end{equation}
At each $t_i$, approximate the empirical target by $\nu_{t_i}=\mathcal{N}(y_i,\varepsilon I)$ with small $\varepsilon>0$ for numerical stability. The empirical risk is
\begin{equation}
\mathcal{L}(t,y,\theta)=\frac{1}{n}\sum_{i=1}^{n}\sum_{k=1}^{K} w_k\, W_2^2\!\Big(\mathcal{N}\big(\mu_k(t_i),\Sigma_k(t_i)\big),\, \mathcal{N}(y_i,\varepsilon I)\Big) \;+\; \lambda \|\theta\|_2^2,
\end{equation}
where $\lambda>0$ is an L2 regularization coefficient. We minimize $\mathcal{L}$ with Adam using mini-batches. For prediction at $t$, the model outputs $\sum_k w_k \mathcal{N}(\mu_k(t),\Sigma_k(t))$; the mean trajectory is $\hat y(t)=\sum_k w_k\mu_k(t)$.

\section{Experiments}
We evaluate on synthetic datasets of varying complexity, including two- and three-dimensional trajectories: \emph{Spiral}, \emph{Ellipse}, \emph{Figure-eight}, \emph{Lissajous figure}, and a 3D \emph{Torus knot}. Baselines are polynomial regression, Gaussian Process Regression (GPR), a Mixture Density Network (MDN), and Wasserstein Barycentric Regression (WBR). We report five metrics: the average Wasserstein distance ($\overline{W_2}$), Energy Distance (ED), negative log-likelihood (NLL), RMSE, and a smoothness index (SRI).

\begin{figure}[H]
    \centering
    \includegraphics[width=0.8\linewidth]{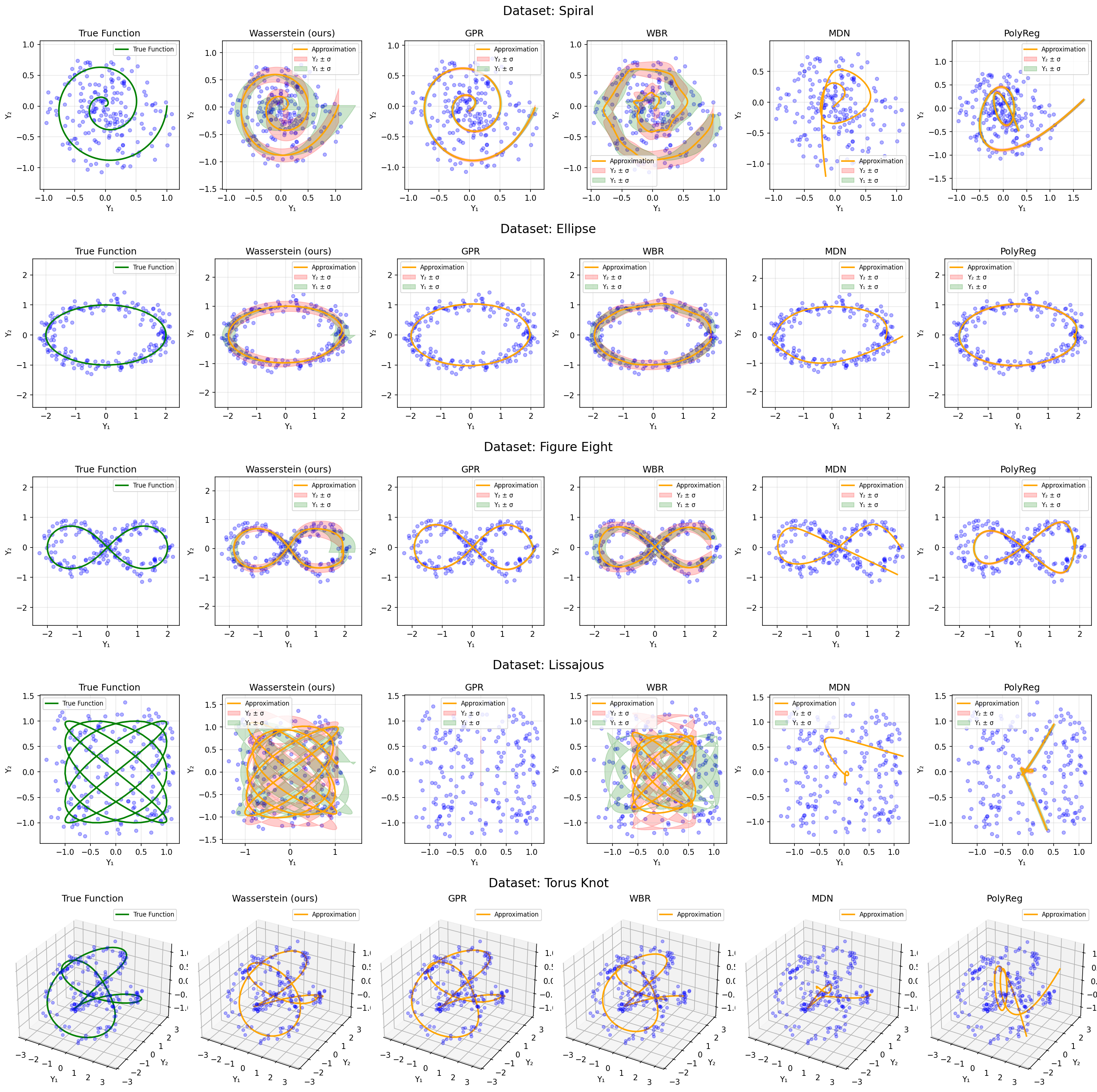}
    \caption{Visualizations of model approximations. From left to right: True Function, Our Wasserstein Regression, Gaussian Process Regression, Wasserstein Barycentric Regression, Mixture Density Network and polynomial regression}
    \label{fig:fig1}
\end{figure}

As soon as we introduce the metrics below (Table~\ref{tab:metrics}), we reference qualitative visualizations in Figure~\ref{fig:fig1}.

\begin{table}[H]
\centering
\caption{Comparison of metrics across models on test datasets. Lower is better for $\overline{W_2}$, ED, and RMSE.}
\label{tab:metrics}
\small
\begin{tabular}{l l c c c c c}
\toprule
Dataset & Model & $\overline{W_2}$ & ED & NLL & RMSE & SRI \\
\midrule
\multirow{5}{*}{Spiral}
& Polynomial  & 0.2477 & 0.1482 & 30.15 & 0.2655 & 0.0020 \\
& GPR         & 0.0559 & 0.0506 & -2.38 & 0.0571 & 0.0014 \\
& MDN         & 0.5901 & 0.2023 & 27964.51 & 0.4840 & 0.0001 \\
& WBR         & 0.2019 & 0.0703 & -1.80 & 0.0687 & 0.0014 \\
& Wasserstein & 0.1678 & 0.0765 & 141.51 & 0.0630 & 0.0015 \\
\midrule
\multirow{5}{*}{Ellipse}
& Polynomial  & 0.0398 & 0.0334 & -3.98 & 0.0467 & 0.0025 \\
& GPR         & 0.0383 & 0.0288 & -4.27 & 0.0368 & 0.0024 \\
& MDN         & 0.1077 & 0.0573 & 4105.04 & 0.0887 & 0.0025 \\
& WBR         & 0.3328 & 0.0611 & -1.13 & 0.0655 & 0.0023 \\
& Wasserstein & 0.2189 & 0.0677 & -1.61 & 0.0559 & 0.0024 \\
\midrule
\multirow{5}{*}{Figure-eight}
& Polynomial  & 0.2383 & 0.1497 & 27.67 & 0.2560 & 0.0028 \\
& GPR         & 0.0557 & 0.0383 & -2.52 & 0.0599 & 0.0028 \\
& MDN         & 0.1953 & 0.0956 & 21461.29 & 0.1940 & 0.0027 \\
& WBR         & 0.3503 & 0.0608 & -0.93 & 0.0925 & 0.0022 \\
& Wasserstein & 0.1398 & 0.0747 & 145.49 & 0.1161 & 0.0028 \\
\midrule
\multirow{5}{*}{Lissajous}
& Polynomial  & 0.9064 & 0.6420 & 456.44 & 0.9612 & 0.0006 \\
& GPR         & 1.3788 & 0.1210 & 2.14 & 1.0000 & 0.0000 \\
& MDN         & 1.2135 & 0.2801 & 71713.58 & 0.8964 & 0.0044 \\
& WBR         & 0.7981 & 0.1185 & 0.38 & 0.3475 & 0.0083 \\
& Wasserstein & 0.2486 & 0.0510 & -1.37 & 0.1091 & 0.0206 \\
\midrule
\multirow{5}{*}{Torus knot (3D)}
& Polynomial  & 1.7320 & 0.6271 & 1687.35 & 1.8421 & 0.0086 \\
& GPR         & 0.0700 & 0.0329 & -4.68 & 0.0735 & 0.0438 \\
& MDN         & 2.5737 & 0.4272 & 179054.0 & 2.0168 & 0.0048 \\
& WBR         & 1.1816 & 0.0986 & 0.27 & 0.4267 & 0.0258 \\
& Wasserstein & 0.4717 & 0.1033 & 0.40 & 0.4031 & 0.0393 \\
\bottomrule
\end{tabular}
\end{table}
\FloatBarrier

Immediately after Table~\ref{tab:metrics}, qualitative visualizations are shown in Figure~\ref{fig:fig1}.

\FloatBarrier

\begin{figure}[H]
    \centering
    \includegraphics[width=0.8\linewidth]{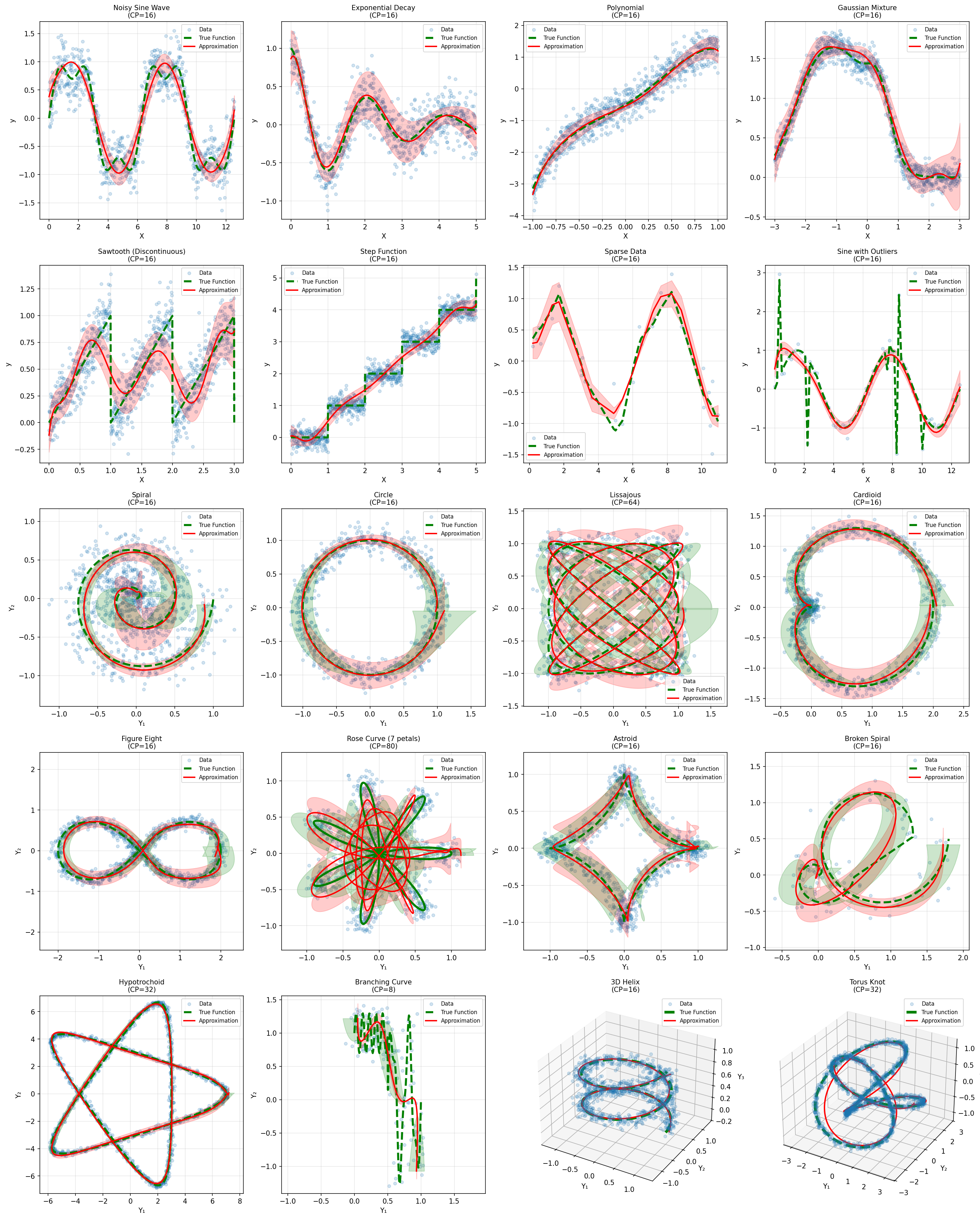}
    \caption{Additional visualizations on extended synthetic functions}
    \label{fig:fig2}
\end{figure}

Additional qualitative results are summarized in Figure~\ref{fig:fig2}.

\FloatBarrier

A comparative analysis shows that the proposed Wasserstein regression method demonstrates stable approximation quality across most tasks, particularly for complex and nonlinear trajectory shapes. Although in some metrics the best results are achieved by GPR and, partially, by WBR, the proposed approach offers fundamental advantages in terms of interpretability and geometrical consistency of distributional representations.

For simple and smooth trajectories, such as the \textit{Ellipse} and \textit{Figure-Eight}, Gaussian processes achieve the lowest RMSE and minimal Wasserstein distance, which can be attributed to their strong local approximation capabilities under moderate noise levels. However, for more complex figures — especially the \textit{Lissajous Curve} — the proposed method outperforms all competitors across the main metrics, providing the smallest values of $\bar{W}_2$ and RMSE, as well as adequate trajectory smoothness (measured by SRI). This indicates that the use of the Bernstein basis combined with Wasserstein distance optimization enables efficient and geometrically consistent modeling of complex trajectory distributions.

\section{Conclusion}
The proposed Bernstein-basis Wasserstein regression provides competitive approximation quality of probabilistic trajectories on synthetic data of varying complexity. The model performs reliably on smooth curves and complex nonlinear trajectories, including three-dimensional structures, as confirmed by Wasserstein distance, Energy Distance, and RMSE. Special attention was paid to geometric correctness of distribution modeling and interpretability of results, achieved through explicit parameterization via control points and the use of the Wasserstein metric. The model maintains a balanced combination of accuracy, trajectory smoothness, and computational stability when trained with standard optimizers. Future work includes moving beyond Gaussian families, applying entropic regularization to improve efficiency, and adapting the approach to multidimensional data for surface approximation and complex structures.

\end{document}